\documentclass[letterpaper, 10 pt, conference]{ieeeconf}  

\IEEEoverridecommandlockouts                              

\usepackage{graphicx} 
\usepackage{booktabs}
\usepackage{multirow}
\usepackage{amsmath} 
\usepackage{bm}
\usepackage{algorithm}
\usepackage{algpseudocode}
\usepackage{lipsum}
\usepackage[table]{xcolor} 

\title{\LARGE \bf
Language-Enhanced Latent Representations \\ for Out-of-Distribution Detection  in Autonomous Driving    
}

\author{Zhenjiang Mao$^{1}$, Dong-You Jhong$^{1}$, Ao Wang$^{1}$, and Ivan Ruchkin$^{1}$
\thanks{$^{1}$Zhenjiang Mao,  Dong-You Jhong, Ao Wang, and Ivan Ruchkin are with the Department of Electrical and Computer Engineering,
        University of Florida, University of Florida, Gainesville, FL, 32603, USA,
        {\tt\small \{z.mao, jhong.dongyou, wang.ao, iruchkin\}@ufl.edu}}%
}

\begin{document}

\maketitle
\thispagestyle{empty}
\pagestyle{empty}

\begin{abstract}



Out-of-distribution (OOD) detection is essential in autonomous driving, to determine when learning-based components encounter unexpected inputs.  Traditional detectors typically use encoder models with fixed settings, thus lacking effective human interaction capabilities. With the rise of large foundation models, multimodal inputs offer the possibility of taking human language as a latent representation, thus enabling language-defined OOD detection. In this paper, we use the cosine similarity of image and text representations encoded by the multimodal model CLIP as a new representation to improve the transparency and controllability of latent encodings used for visual anomaly detection. We compare our approach with existing pre-trained encoders that can only produce latent representations that are meaningless from the user's standpoint. Our experiments on realistic driving data show that the language-based latent representation performs better than the traditional representation of the vision encoder and helps improve the detection performance when combined with standard representations.


\end{abstract}

\section{INTRODUCTION}

Human-machine cooperation often requires dynamic and intricate interactions between the driver (human) and the vehicle's autonomy (machine), especially in the context of autonomous driving. These interactions require not only the physical control of the vehicle --- but also carefully considering how drivers understand, trust, and effectively communicate with the autonomous systems~\cite{koopman2022safe,Cosner2023LearningResponsibility}. As vehicles become increasingly autonomous, the interface between humans and machines evolves, requiring intuitive and transparent communication channels~\cite{Liu2021ComputingSystems}. For example, this interface must provide clear, timely, and actionable information, allowing drivers to swiftly comprehend the autonomous system's decisions and actions. The effectiveness of human-machine interaction in driving is critical to ensuring that the transition between manual and autonomous control is seamless and safe~\cite{Lee2015AutonomousVehicles,9629359}. On the flip side, the driver needs expressive ways of instructing the vehicle and limiting its autonomy. The process entails a continuous exchange of feedback, where the driver's actions and the autonomous system's reactions are synchronized to ensure the highest level of driving performance and safety. Thus, enhancing human-machine interaction in autonomous driving systems is pivotal in building driver trust, ensuring safety, and facilitating the acceptance and adoption of autonomous vehicles in society.

In autonomous driving, the concept of \textit{latent representations} plays a key role in enhancing the system's performance and reliability~\cite{li2019conditional,wen2022social,10342070}. By converting high-dimensional image data into a more manageable low-dimensional latent space, these representations significantly reduce computational complexity and speed up processing, which is crucial for real-time decision-making. Latent representations abstract the essential features of the data, distilling the information to its most relevant aspects for the task at hand~\cite{wang2019deepvid,8667661}. One key task accomplished with latent representations is the quick identification of anomalies and out-of-distribution samples~\cite{salehi2021unified}. Another common use case is improving the generalization from the training data to new, unseen scenarios~\cite{9782500,volpi2018generalizing}. Consequently, leveraging latent representations not only bolsters the system's performance in standard operational settings --- but also enhances its robustness and adaptability to diverse and unforeseen driving conditions, thereby leading to more trustworthy and dependable autonomous driving.


\begin{figure}[t!]
  \centering
  \includegraphics[width=0.9\linewidth]{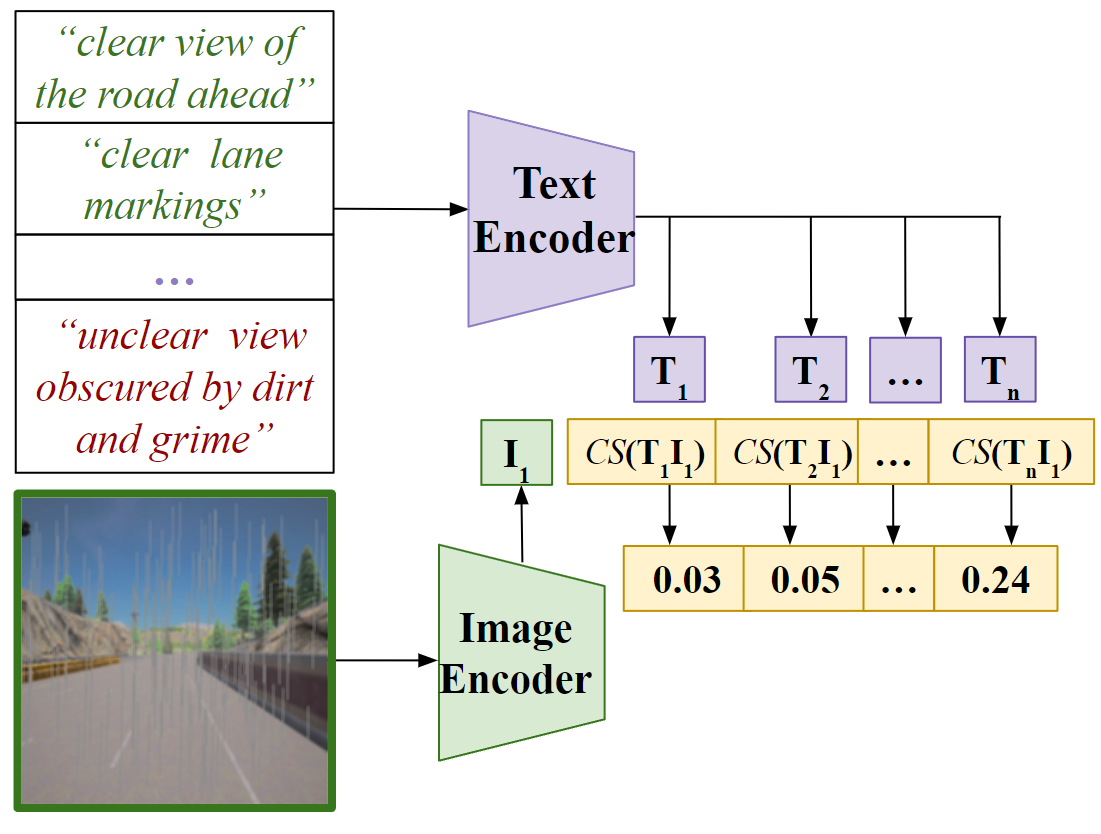}
  \vspace{-3mm}
  \caption{Examples of language-based encoding with normal (green texts) and anomalous description (red texts) by calculating the \textit{cosine similarity (CS)} between the text and image representations.}
    \label{fig:problem}
  \vspace{-6mm}
\end{figure}

\looseness=-1
Successful training of a latent-based anomaly detector requires \textit{sufficient and high-quality training data}. Over the past decade, this training typically required often expensive tasks: data collection, augmentation, targeted sampling, and tuning the performance, generalization, and robustness of learning components before deployment~\cite{hendrycks2021many}. 
As pre-trained models such as ResNet~\cite{he2015deep} and Visual Transformer (ViT)~\cite{dosovitskiy2021image,raghu2021vision} trained on large datasets like ImageNet~\cite{5206848} become more popular, they can be reused for powerful visual feature extraction capabilities \textit{without} painstakingly obtaining and curating large amounts of training data. For instance, ViT's attention mechanism can improve recognition accuracy by allowing the model to focus better on special parts of the image~\cite{radford2021learning}, leading to the improved reliability and performance of vision-based tasks such as anomaly detection.
Bypassing the training via pretrained models offers a significant speed up to the development --- and thus faster iterations leading to early adoption of autonomous driving. However, when reusing pretrained models, the tuning of anomaly detection would still be done before deployment and without any input from the final user (the vehicle's driver or passenger), hence inhibiting the transparency and trustworthiness of the autonomy.

This paper explores a novel paradigm of \textit{letting users set the focus of anomaly detection} on phenomena that they care about described in natural language, in the context of the detection of out-of-distribution (OOD) for autonomous driving~\cite{Yi2021ImprovedOOD,9857153,quinonero2008dataset}. We term this novel capability \textit{language-enhanced latent representations}: it allows the user to prompt the OOD detector with the nominal distribution of interest as shown in Fig~\ref{fig:problem}. For example, the driver can specify that they expect the car to see a clear, bright, and open road --- and everything else should be treated as an OOD input. We hope that this capability drastically increases the flexibility and transparency of anomaly detection from the standpoint of end users. Our implementation of this idea on pretrained models shows promising results in experiments on distribution shifts of camera data from a photorealistic driving simulator.


In this paper, our contributions are as follows:
\begin{enumerate}
    \item A new language-guided OOD detection technique that gives the end user more transparency and control.
    \item Extensive experiments on photorealistic simulated data to investigate the performance of different language encodings for OOD detection.
    
    
    
    

\end{enumerate}

\section{Related Work}

\subsection{OOD Detection and Risk Assessment}

In the domain of autonomous driving, safety and performance are crucially dependent on the ability to effectively detect out-of-distribution (OOD) data that could lead to significant performance degradation~\cite{Yi2021ImprovedOOD,9857153,quinonero2008dataset}. To mitigate the risks posed by OOD and other unexpected inputs, there has been a development of advanced algorithms and safety mechanisms~\cite{bitterwolf2020certifiably,ruchkin2022confidence}. These utilize techniques such as image-based analysis and conversion to low-dimensional latent representations to enhance efficiency and reduce computational load. Moreover, the field of anomaly detection has evolved significantly, employing temporal analysis, varied methodologies, and sophisticated models like autoencoders and generative adversarial networks to improve safety assessments~\cite{xia2022gan,stocco2019misbehaviour,goodfellow2014generative}. One multidimensional approach integrates formal specifications with anomaly detection, enhancing safety prediction and generating actionable scores that improve the reliability and safety of autonomous driving systems~\cite{sekar2002specification,fauri2017system}. However, all these approaches have not considered direct English-language inputs from users to define what constitutes an OOD input. 

\subsection{Latent Representations}

There have been significant advancements in large deep neural networks and transformers, alongside their respective latent representations~\cite{guo2022cmt,raghu2021vision}. These models employ latent vectors to encode the crucial features of input data into a more concise and computationally tractable  form~\cite{flamich2020compressing,hedayati2022model}. This strategy not only reduces the computational complexity but also bolsters the model’s ability to generalize from seen to unseen data, thereby enhancing its performance across various tasks and contexts. The synergy between large model architectures and latent representation techniques has led to significant advancements in the field~\cite{flamich2020compressing}. This combination offers robust and scalable solutions for complex problems like OOD detection in autonomous driving systems, marking a pivotal development in ensuring the safety and reliability of these systems~\cite{9857153}.

\subsection{Multimodal Models}

\begin{figure*}[tbh]
\centering\includegraphics[width=1.0\linewidth]{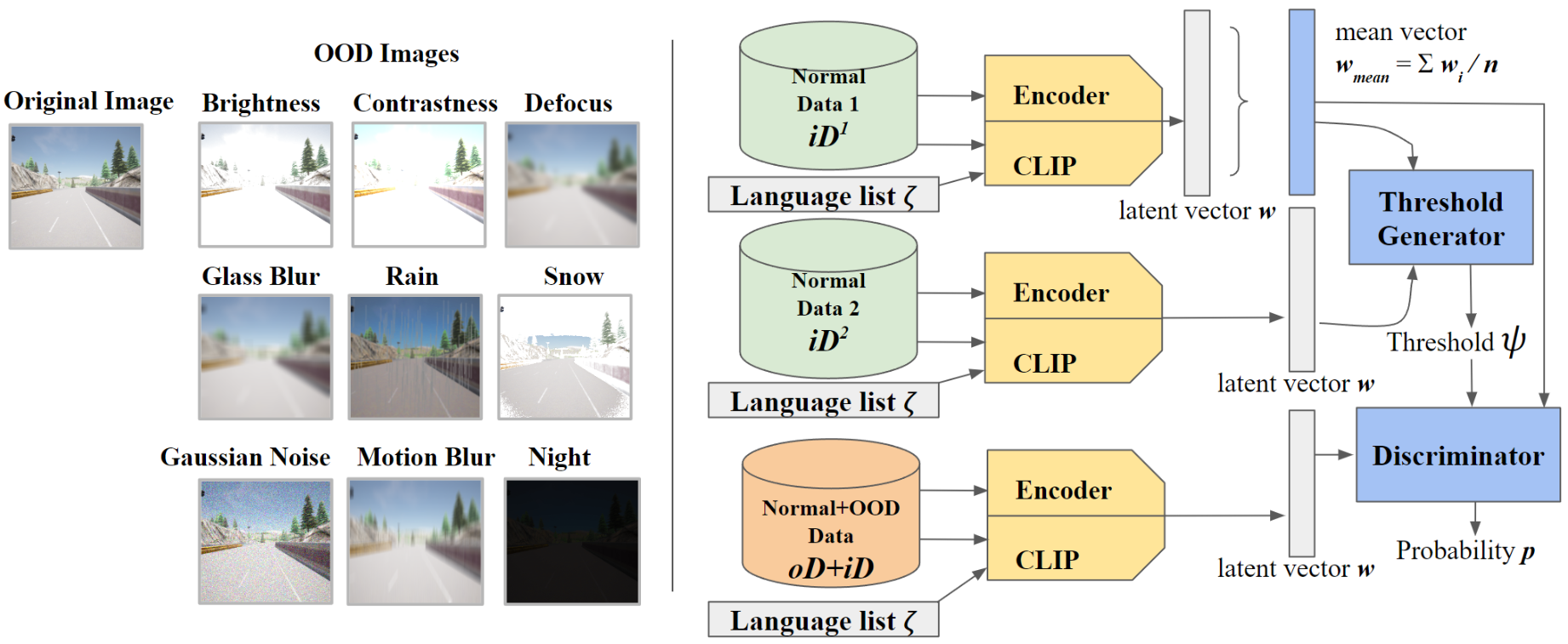}
  \caption{Out-of-distribution types (left) and the architecture of OOD detection with latent representation distance (right).}
    \label{fig:overview}
\end{figure*}

In the realm of multimodal machine learning, significant strides have been made in integrating diverse modalities such as text, images, and audio to enhance machine understanding~\cite{atrey2010multimodal}. 
Recent advancements in deep multimodal learning~\cite{baltruvsaitis2018multimodal} have been made like the impact of neural architectures on joint representation learning from visual and linguistic inputs~\cite{ramachandram2017deep}. Enhancing the alignment of and translation between modalities is another important consideration. For example, ViLBERT~\cite{lu2019vilbert} utilizes a dual-stream transformer architecture to independently process visual and textual streams before interaction, which has proven effective in tasks requiring fine-grained cross-modal understanding. Similarly, the emergence of models like CLIP showcases the power of contrastive learning in bridging the semantic gap between text and images, performing well in zero-shot tasks by aligning textual and visual representations in a shared embedding space~\cite{radford2021learning}. 
Our paper exploits the ability of multimodal approaches to bridge the gap between latent representations of different types of input, exploiting a connection between in-/out-of-distribution images and user-provided textual descriptions. 

\section{Preliminaries and Problem Statement}

As automated driving gradually becomes a mainstream technology, most vehicles are equipped with software-based safety assistant systems to help the driver avoid misbehavior and even perform fully automated driving based on sensor data. It is the best practice for such learning-based systems to come with supervisory components that can detect misbehavior and stabilize/safeguard the system (e.g., by stopping or pulling over)~\cite{phan2020neural,10.1007/s10515-021-00310-0,Hobbs_2023,gleirscher2023supervision}. A key element of such supervision is the detection of unexpected inputs to learning-based components. We will specifically focus on OOD detection for images because camera data is among the most common. 

\looseness=-1
Suppose an image-based neural network system $s$ takes camera images $\bm y_t$ at time $t$ as input and has an output $u_t$. For example, this system can be a controller learned by reinforcement learning like Deep Q Learning~\cite{vanhasselt2015deep} to determine the action for the next moment or a safety predictor~\cite{mao2024safe} to predict if the car will suffer from any potential danger in the next few moments given an image observation $\bm y_t$ of the current moment.

\section{OOD Detection Approach}

The self-driving system $s$ is usually trained by supervised learning methods and the training data is called in-distribution data ($iD$) relative to OOD data ($oD$). System $s$ has some form of assurance of performing well on $iD$ (e.g., due to formal verification~\cite{chen2022reachability,geng2024bridging} or statistical bounds~\cite{qin2022statistical}), but has no certainty regarding its performance on $oD$. Our general problem is defined as follows: given any input $\bm y$, an OOD detector's goal is to determine whether this image $\bm y$ belongs to $iD$ or $oD$.

\subsection{Language-enhanced representation}
In traditional OOD detections~\cite{stocco2019misbehaviour}, images will be compressed into latent representations and detect how much new observed images deviated from the given known images by calculating the distance of latent representations. However, these representations encoded from models like VAE are not interpretable, and the calculation of distribution distance on each value or to say each dimension of latent representations is not transparent like trivial dimensions are accounted for a lot but neglect to add more weights on important ones, which means methods based on meaningless encoding are not trustworthy. Casual representations~\cite{yang2023causalvae} require additional labels and unsupervised interpretable encoding is computationally intensive~\cite{mao2024zeroshot}.   Instead of calculating the reconstruction loss from decoder~\cite{stocco2019misbehaviour}, we need to compute a mean vector and calculate the distance distribution of distances between known data and the mean vector. First, the in-distribution dataset  $\bm Y_{iD}$ is split into two parts after shuffling: $\bm Y_{iD}^v$ and  $\bm Y_{iD}^f$ for calculating the mean vector and fitting the distance distribution. The encoder $E(\bm y)=\bm \omega$ compresses all data of $\bm Y_{iD}^v$:  $[\bm y_1^v, \bm y_2^v,..., \bm y_n^v]$ into latent representations $\bm V= [\bm v_1^v, \bm v_2^v,..., \bm v_n^v]$. We also adopt the image encoder of multimodal model CLIP~\cite{radford2021learning} to compress the images into another representation: $[\bm q_1^v, \bm q_2^v,..., \bm q_n^v]$. Then we have a sentence set $\zeta$ with sentences describing the images coming up by humans, where Fig.~\ref{fig:problem} exemplifies such sentences with a mixed normal and anomalous description of driving situations. The text encoder of CLIP compresses $\zeta$ into latent representations: $\bm K = [\bm \kappa_1^v, \bm \kappa_2^v, ..., \bm \kappa_m^v]$. The cosine similarity of each $\bm q$ and $\bm \kappa$ will be calculated to form a language-based representation:
\begin{equation}
     \pi  = \frac{{\bm q \cdot\bm \kappa}}{{\|\bm q\| \cdot \|\bm \kappa\|}}
\end{equation}

After getting a set for all $\bm y$ in $\bm Y_{iD}^1$: $\{\bm \pi_1^v, \bm \pi_2^v,..., \bm \pi_n^v\}$, these language-based representations can act as latent representation alone and also added or appended into origin encoded the original latent $\bm V$ to get another novel type of representation: $[\bm \omega_1^v, \bm \omega_2^v,..., \bm \omega_n^v]$. 

\begin{table*}[th!]
\centering
\begin{tabular}{llllllllllll|lllllll}
        \toprule

\multirow{3}{*}{Model} & \multirow{3}{*}{$\omega$ type} & \multicolumn{12}{c}{Comparison} \\ & & \multicolumn{6}{c}{}   \\

        \midrule
            &        & rain     & snow        & night     & bright & fog        & contrast     & defocus     & gauss        & glass     &motion & 
            \textbf{mean}   &  \textbf{std}      \\ 
            \hline

  \textit{SelfOracle~\cite{stocco2019misbehaviour}}   & $v_{}$& {88.03} & 88.68 & 60.06 & 88.84 & 60.56 & \textbf{89.08} & 59.81 & 81.77 & 59.88 & 60.07 & 73.67 & 13.74
  
\\

\hline
   VAE & $v_{}$ & 62.85 & 79.94 & 78.84 & 64.77 & 63.02 & 63.74 & 63.82 & 63.58 & 63.83 & 63.84 & 66.82 & {6.307}

   \\

     ResNet & $v_{}$ & 66.64 & 78.30 & \textbf{95.56} & 66.62 & {82.01} & 79.09 & 95.76 & 69.43 & 95.76 & {89.33} & 81.85 & 11.26

\\
  ViT & $v_{}$ & 70.14 & 66.89 & 67.59 & 65.44 & 75.67 & 68.41 & 87.30 & 64.69 & 95.60 & 71.86 & 73.35 & 9.718


\\
      \textit{LLR} (Normal desc.) & $\pi_{}$ & 63.96 & 83.46 & 87.82 & \textbf{90.31} & 66.22 & 82.18 & 65.23 & 69.98 & 64.58 & 64.45 & 73.81 & 10.23
\\
      \textit{LLR} (Anomalous desc.)&$\bar{\pi}_{}$ & 63.06 & 68.74 & 92.50 & 82.66 & 79.34 & 84.13 & 91.11 & 62.47 & 82.17 & 62.84 & 76.90 & 11.08
\\
      \textit{LLR} (Mixed desc.)&$( \pi_{},\bar{\pi}_{})$ & 64.20 & 83.05 & 91.44 & 85.48 & 67.57 & 76.45 & 66.18 & 66.65 & 65.60 & 64.66 & 73.12 & 9.626
\\
        \textit{LLR} (v added)&$\pi_{}+v_{} $ & 87.19 & 72.82 & 78.46 & 65.04 & 75.73 & 76.91 & 90.59 & 74.60 & 90.49 & 71.98 & 78.38 & 8.041
\\
       \textit{LLR} (v added)&$\bar{\pi}_{}+v_{} $ & 66.16 & 67.57 & 88.19 & 65.42 & 65.48 & 65.62 & 89.68 & 65.67 & 75.01 & 68.49 & 71.72 & 9.031
\\
   \textit{LLR} (v appended)&$(\pi_{},v_{})$ & 72.14 & 69.92 & 75.06 & 65.44 & 77.47 & 67.09 & 78.55 & 68.11 & 80.20 & 70.10 & 72.40 & \textbf{4.878}
\\
\textit{LLR} (v appended)&$(\bar{\pi},1v_{})$ & 69.01 & 66.87 & 93.86 & 71.81 & 72.12 & 66.74 & 92.05 & 64.35 & 94.07 & 76.73 & 76.76 & 11.33
\\
\textit{LLR} (v appended)&$(\pi_{},2v_{})$ & 67.61 & 71.77 & 69.24 & 65.05 & 86.93 & 72.40 & 93.92 & 66.85 & 84.69 & 82.54 & 76.10 & 9.535
\\
\textit{LLR} (v appended)&$(\bar{\pi},2v_{})$ & 64.55 & 65.18 & 74.15 & 66.94 & 87.49 & 67.77 & 81.33 & 64.88 & 95.25 & 69.80 & 73.73 & 10.21
\\

\textit{LLR} (v appended)&$(\pi_{},3v_{})$ & 73.49 & 68.82 & 69.24 & 67.75 & 71.00 & 76.64 & 94.21 & 65.55 & 94.15 & 72.33 & 75.31 & 9.879
\\
\textit{LLR} (v appended)&$(\bar{\pi},3v_{})$ & 71.91 & 68.48 & 68.51 & 69.30 & 72.34 & 67.21 & 88.94 & 72.71 & 94.87 & 76.79 & 75.10 & 8.898
\\
\textit{LLR} (v appended)&$(\pi_{},4v_{})$ &\textbf{88.87 }& 68.82 & 75.39 & 66.68 & \textbf{89.14} & 72.62 & 90.55 & 65.30 & 95.43 & 71.97 & 78.47 & 10.71
\\
\textit{LLR} (v appended)&$(\bar{\pi},4v_{})$ & 64.95 & 69.78 & 69.28 & 69.61 & 70.44 & 67.04 & \textbf{95.79} & 64.66 & 82.56 & 77.17 & 73.12 & 9.165
\\
\textit{LLR} (v appended)&$(\pi_{},5v_{})$  & 75.04 & 71.99 & 74.34 & 68.16 & 81.19 & 68.33 & 95.61 & 65.81 & 93.67 & 71.29 & 76.54 & 9.933
\\
\textit{LLR} (v appended)&$(\bar{\pi},5v_{})$ & 68.86 & 68.59 & 74.60 & 67.05 & 84.49 & 67.72 & 93.49 & 65.48 & 95.01 & 73.06 & 75.83 & 10.55

\\
\textit{LLR} (Normal desc.)& $2\pi_{}$& 66.64 & \textbf{92.07} & 65.26 & 86.90 & 67.75 & 78.80 & 88.71 & 75.12 & 80.95 & 80.22 & 78.24 & 9.001

\\
\textit{LLR} (Anomalous desc.)&$2\bar{\pi}_{}$ & 67.14 & 88.27 & 94.43 & 76.43 & 69.17 & 84.93 & 88.54 & \textbf{93.95} & 77.78 & \textbf{95.01} & \textbf{83.56} & 9.833

\\

   \textit{LLR} (Mixed desc.)&$(2\pi_{},2\bar{\pi}_{})$& 63.87 & 80.74 & 92.66 & {90.21} & 81.95 & 87.21 & 88.67 & 66.60 & 80.92 & 65.17 & 79.79 & 10.27

  \\

       \textit{LLR} (v added)&$2\pi_{}+v_{} $& 77.75 & 73.58 & 68.27 & 66.48 & 81.38 & 72.76 & 81.13 & 65.24 & 95.54 & 67.76 & 74.98 & 8.833
    \\
   \textit{LLR} (v appended)&$(2\pi_{},v_{})$ & 70.64 & 68.61 & 66.49 & 67.74 & 74.18 & 69.85 & 88.92 & 64.39 & 79.24 & 69.08 & 71.91 & 6.901
   \\
     \textit{LLR} (v appended)&$(2\pi_{},2v_{})$ & 71.04 & 68.34 & 66.79 & 66.42 & 79.32 & 70.24 & 92.25 & 65.47 & 81.11 & 69.02 & 73.00 & 8.147
     \\

       \textit{LLR} (v appended)&$(2\pi_{},3v_{})$ & 80.02 & 71.67 & 69.59 & 67.18 & {85.25} & 68.02 & {95.63} & 65.76 & 94.39 & 86.81 & 78.43 & 10.90
       
       \\ 
       \textit{LLR} (v appended)&$(2\pi_{},4v_{})$ & 71.29 & 67.84 & 65.66 & 66.64 & 82.72 & 67.20 & 81.21 & 65.85 & \textbf{95.79} & 75.26 & 73.94 & 9.413
      \\

       \textit{LLR} (v appended)&$(2\pi_{},5v_{})$ & 67.99 & 69.85 & 68.01 & 66.34 & 75.31 & 68.40 & 92.79 & 64.91 & 95.05 & 83.29 & 75.19 & 10.63
\\

        \bottomrule

\end{tabular}
\caption{Comparison of F1 scores for $oD$ detection. Parentheses and plus symbols stand for the operations of appending and adding.}
\label{tab:f1}
\end{table*}

\subsection{Threshold Estimation}

We calculate a deterministic mean of this latent vector: 
\begin{equation}
    \bm \omega_{mean} = \frac{\Sigma^n_{i=1} \bm \omega_i^v}{n}
\end{equation}

Next, by compressing images $[\bm y_1^f, \bm y_2^f,..., \bm y_n^f]$ from $\bm Y_{iD}^f$ into latent space  with the same procedure as for $\bm Y_{iD}^f$: $ [\bm \omega_1^f, \bm \omega_2^f,..., \bm \omega_n^f]$, we obtain the distance:
\begin{equation}
\bm \epsilon = \lambda(\bm \omega, \bm \omega_{mean})
\end{equation}
between each latent vector from $\bm Y_{iD}^f$ and the $\bm \omega_{mean}$ can be computed by a distance metric $\lambda$, for which we use cosine similarity (CS) in this paper. The distance set  $\{ \epsilon_1,  \epsilon_2,...,  \epsilon_n\}$ is fitted into a Gamma distribution $\Gamma$ inspired by the work of \textit{SelfOracle}~\cite{stocco2019misbehaviour} which is a baseline in our experiment. 



After, we fit the distance set $\bm \epsilon$ into particular one-dimensional distributions which is the Gamma distribution here: $\Gamma$, we set a confidence level $\varphi$ (e.g.,  $90\%$) to be the threshold point. Any data with a distance below this point can be considered as OOD data. To threshold the normal vs anomalous data, we use the cumulative distribution function $F_\Gamma$ of $\Gamma$ to determine the threshold $\psi = F_\Gamma^{-1}(\varphi) $, below which are normal points and above which are anomalous. 




 In order to compare the performance of different vision models in processing images related to autonomous driving, we have adopted three representative models: (i) the encoder of a VAE (a Convolutional Neural Network, CNN with 2 convolutional layers), (ii) ResNet-50, and (iii) a Vision Transformer (ViT). Only the VAE needs to be trained on the $\bm Y_{iD}$, while the other two models are pre-trained on ImageNet~\cite{5206848} and do not have access to \textit{any} normal and OOD data.

\section{Results}

In our experiments, CARLA~\cite{dosovitskiy2017carla} was used to simulate a town environment containing up to 20 self-driving vehicles driving on the road at the same time. Such a setup simulates real-world town driving conditions, including multi-vehicle interactions and various road conditions. All normal data is collected in a setting of a clear noon town road and we changed the weather and sun position of the internal setting of the CARLA system and conducted various scripts to generate different kinds of OOD data.
To balance the precision and recall, we choose F1 score to display the results: $F1 = 2 \times \frac{\text{Precision} \times \text{Recall}}{\text{Precision} + \text{Recall}} = 2 \times \frac{TP}{{TP + \frac{FP + FN}{2}}}$. The test data is composed of a 1:1 mix of in-distribution and out-of-distribution data.

\looseness=-1
Table~\ref{tab:f1} shows the performance of different model and representation $\omega$ types which conclude pure normal representation, pure language-based representation, and composite representations with languages of normal, anomalous, and mixture sentences. The $\omega$ in the table indicates the type of representations: $v$ stands for the pure normal representation from an encoder, $\pi$ stands for the pure language-based representation, and $\bar \pi$ stands for the anomalous sentences of language description. For example, $(\pi,3v)$ denotes appending the language-based representation with a three-times-length normal representation. Our proposed method Language Lantent Representation is abbreviated as \textit{LLR}.

\looseness=-1
ResNet is found to be the best model for pure normal representation methods, and it performs well on various OOD types. Adding a language-based representation, the F1 score increases merely, and the appending operation dramatically enhances the F1 score. Moreover, we observed that the best length ratio of language-based representation to the normal representation is 1:3. In pure language-based representations, the language of anomalous description performs better than the other two of normal and mixed description, which is the best method.

\section{Conclusion and Discussion}
Our research explores a new approach to anomaly detection in the autonomous driving domain, called language-augmented latent representation. We introduce an OOD detection technique that revolutionizes anomaly detection methods by leveraging language-augmented latent representations. This innovative paradigm enables users to focus their anomaly detection efforts on specific phenomena of interest expressed in natural language. With this feature, drivers can specify expectations for scenarios such as clear, bright, and open roads, indicating that any deviations should be flagged as out-of-distribution (OOD) inputs. Furthermore, we conduct extensive experiments on realistic simulation data to identify pre-trained models that perform well in OOD detection for autonomous driving applications, laying the foundation for practical implementation. We anticipate that
this approach will not only enhance anomaly detection accuracy but also provide users with transparency and language-based control over the process.

Our analysis provides insight into the performance of different methods under various image distortion scenarios, demonstrating the robustness of the \textit{SelfOracle} method~\cite{stocco2019misbehaviour} and the promising and promising performance of the \textit{LLR} method. This is critical for identifying out-of-distribution images and fostering human trust in autonomous systems. We also highlight the required balance between average accuracy and adaptivity to achieve optimal model performance.

Our results show that our proposed language-enhanced representation OOD detector performs well in detecting OOD images using anomalous descriptions as inputs of text encoder, achieving the highest accuracy and F1 score among the models. This highlights the potential of these models in autonomous driving systems and the need for future research to explore model adaptability and optimization of the natural language conversion process. 

\looseness=-1
Due to the use of the same textual prompts for all test situations, the result of pure language representation shows the best average F1 scores but underperforms in some specific situations like rainy scenes. To improve scene-specific performance, different sets of prompts can be developed in the future. Additional techniques can also be developed to use image-text models to \textit{adjust} the prompts and adapt to users' \textit{interactive} instructions to further optimize the representations.

\bibliographystyle{IEEEtran}

\bibliography{sample}

\end{document}